\pdfoutput=1
\documentclass[11pt]{article}
\usepackage{ACL2023}
\usepackage{times}
\usepackage{latexsym}
\usepackage[T1]{fontenc}
\usepackage[utf8]{inputenc}
\usepackage{microtype}
\usepackage{inconsolata}
\usepackage{xspace}

\definecolor{CMpurple}{rgb}{0.6,0.18,0.64}

\newcommand\eg{e.g.\xspace}
\newcommand\ie{i.e.\xspace}
\newcommand\cf{cf.\xspace}

\title{Evaluation for Change}
\author{Rishi Bommasani \\
  Stanford University \\
  \texttt{nlprishi@stanford.edu} 
}

\begin{document}
\maketitle

\begin{abstract}
Evaluation is the central means for assessing, understanding, and communicating about NLP models.
In this position paper, we argue evaluation should be more than that: it is a force for driving change, carrying a sociological and political character beyond its technical dimensions.
As a force, evaluation's power arises from its \textit{adoption}: under our view, evaluation succeeds when it achieves the desired change in the field.
Further, by framing evaluation as a force, we consider how it competes with other forces.
Under our analysis, we conjecture that the current trajectory of NLP suggests evaluation's power is \textit{waning}, in spite of its potential for realizing more \textit{pluralistic} ambitions in the field. 
We conclude by discussing the legitimacy of this power, who acquires this power and how it distributes.
Ultimately, we hope the research community will more aggressively harness evaluation for change.
\end{abstract}
\section{Introduction}
\label{sec:introduction}

Evaluation plays a defining role in NLP research; in fact, evaluation has a very rich history.
While this genealogy can be traced in many ways, since this piece (roughly) coincides with the 5\textsuperscript{th} anniversary of the passing of one NLP's beloved pioneers and the first recipient of the ACL Lifetime Achievement Award, we look to Aravind Joshi's legacy.
Best known for grammar formalism and discourse \citep[see][]{webber2018obituary}, his research journey reflects broader field-wide trends towards evaluation.
In early works \citep[\eg][]{joshi1969sag, joshi1972sag, grosz1983providing}, evaluation went entirely unmentioned.
Yet, over time, Aravind's work involved more evaluation \citep[\eg][]{joshi1989evaluation}, implicitly building new norms for evaluation in grammar formalism and discourse \citep{miltsakaki2004pdtb, prasad2008pdtb2, prasad2014reflections}.
\citet{liberman2005joshi} cites Joshi's standards for evaluation in conveying Joshi's signature belief in multidisciplinary approaches to human language.

Joshi's life and 5 decades of scholarship teaches us evaluation is important, but how should we reason about evaluation?
Here, we present two perspectives that frame evaluation in considerably different ways.
Under the first account, evaluation is technical in nature, functioning as a lens to study models.
The motivation for this lens may depend on the specific evaluation, stakeholder, or both:
evaluation may allow us to derive scientific insight.
Or it can transparently document technology for broader audiences (\eg practitioners, colleagues in other fields, policymakers, the public). 
Regardless, to determine if an evaluation is successful, under this account, the lens must yield the desired understanding about models.

In this work, we argue for a second perspective, which we believe is partially acknowledged but considerably less salient than the first perspective.
Under our second account, evaluation is political in nature, functioning as a force to drive change.
In contrast to the first account, this means evaluation pushes the research community in some direction, possibly referring to a specific social or scientific objective, with the emphasis being on future model development more so than existing models.
Critically, under this account, to determine if an evaluation is successful, the force must yield the desired change in the community.
By separating these two accounts, our goal is neither to suggest they are at odds nor that they are meaningfully separable, but to shed conceptual clarity on the merits of power-centric analysis.

In pushing for this position of viewing evaluation as a force, we explore what this force influences, what other forces it competes with, how it accrues power, whether its power is legitimate, and who it empowers.
Motivated by the growing impact of language technology and our field, the abundant discord on the status quo, and the uncertainty on what lies ahead, we believe evaluation's potential for change presents a vital path forward.
\section{Evaluation as a Force}
\label{sec:power}
If evaluation is a force, what domain does it act upon?
And where does its power come from?

\paragraph{Domain.}
We will restrict our scope to how evaluation influences NLP research.
Specifically, evaluation concretizes desired behavior for systems, thereby communicating an objective for model design.
This allows for the community to coordinate on goals for modeling research.
For this goal-setting to succeed, future research should then go on to make progress on the proposed evaluation.
That is, successful evaluation requires that the evaluation be prioritized, redistributing research attention such that it is allocated towards making progress on the evaluation.

\paragraph{Adoption constructs power.}
As this suggests, the adoption of an evaluation (by others) generates its power and determines its success.
It is in this sense that our account for evaluation success deviates from a purely technical/intrinsic characterization.
Most evaluations are concrete instantiations of a broader agenda: for these evaluations to be effective, they must shift power, namely towards addressing this agenda and materially making progress. 
In spite of this, we generally find that evaluations in NLP research do not even mention how adoption will arise, and if evaluation creators will take any overt actions to accelerate adoption.

\paragraph{Accelerating adoption.}
If the power of evaluations come from adoption, and evaluation creators are incentivized to accrue such power to advance their broader agenda, are there ways to accelerate adoption?
We observe at least two such approaches, though they have not been considered in this way to our knowledge.
As a softer means for acquiring adoption/power, evaluations may be used as shared tasks \citep[\eg~SemEval; see][]{parra2017shared, nissim2017sharing} or be built as part of workshops/conferences \citep[\eg~BIG-bench; see][]{welm2021welm, srivastava2022bigbench}, which leans into the relationship between coordinating research and convening researchers.
More aggressively, explicit competitions with prizes or other stronger incentives can more directly drive adoption, perhaps most famously in the Netflix Prize, which remarkably accelerated and shifted research on recommender systems \citep[see][]{hallinan2016netflix}. 

\paragraph{Authority as a standard.}
As evaluations accrue influence, they eventually become reified as high-status standards like ImageNet, WMT, and SQuAD \citep{dotan2020value, dehghani2021lottery}. 
While it is difficult to directly assess the power these evaluations have \citep[\eg how would research have changed counterfactually in their absence; see~][]{liu2021small}, strong norms emerge for modeling work to evaluate on these standards.
And, consequently, improvements on these evaluations function as stand-ins for more fundamental progress \citep{liao2021are, raji2021ai}.
In fact, their authority is made clear in how serious improvements were seen as watershed moments, ushering in new paradigms.
Famous examples include the performance of AlexNet \citep{krizhevsky2012alexnet} on ImageNet, which initiated the deep learning revolution, and Transformers \citep{vaswani2017transformers} on WMT, which, by outperforming specialized machine translation approaches by a considerable margin, marked the dawn of the current dominance of Transformers. 

\paragraph{Related work.}
This work is not the first to bring questions of power, values, reflection, and change to the fore in relation to evaluation/benchmarking 
\citep[][\textit{inter alia}]{
jones1995evaluating, 
welty2019metrology,
dotan2020value,
ethayarajh2020utility, 
linzen2020accelerate,
scheuerman2021politics,
kiela2021dynabench, 
dehghani2021lottery,
bowman2021fix, 
raji2021ai,
koch2021reduced,
denton2021genealogy,
paullada2021data,
liu2021small,
hutchinson2021towards,
jacobs2021measurement,
birhane2022values,
liang2022helm}.
Prior work establishes that evaluations embed values, carry influence, encode broader power structures, and the nature of evaluation as ranking aligns with broader themes of hierarchy.
They make clear how other disciplines can provide guidance on what we see in NLP, but also how our evaluation practices are distinctive (\eg competitive tendencies in in benchmarking, differences in standards for measure validity).

While we draw significant inspiration from these works, our work also significantly diverges in its objective.
Rather than trying to make visible the tacit assumptions, norms, and infrastructure that animate evaluation's power, we instead set our sights on how evaluation's power can animate change.
In this regard, our work more closely mirrors the aesthetic of \citet{abebe2020roles}, as can be seen in the similar titles.
\section{Competing Forces}
\label{sec:competing-forces}

Having argued for where evaluation draws power from, how powerful is it?
While difficult to state in absolute terms, we instead consider what other forces are in play and how they interact/compete.

\paragraph{Coexisting forces.}
NLP research is a fabric stitched through myriad social interactions: 
conversations with colleagues, talks at conferences, academic Twitter, scholarship from adjacent disciplines, and much more.
In our view, most of these interactions are poorly conceptualized as forces: 
while they exert influence, they are generally diffuse rather than concentrated and lack strong directionality. 
For this short-form analysis, we juxtapose evaluation with the force of \textit{resources}.
By resources, we refer to assets like money, compute, and engineering support, choosing to treat them as monolithic (rather than disaggregating) for this short-form analysis.

\paragraph{Language models.}
Given the central position language models occupy in modern NLP, we take language models as a case study to consider the forces of evaluation and resources.
Our thesis is resources, to a far greater extent than evaluation, govern research on language models, with second-order effects for broader NLP research given the pervasive dependence on language models in the field at present.
Beyond the fact that influential language models have near-exclusively been developed by resource-rich institutions, resource-allocation mindsets appear to drive decision-making in  their development.
Namely, the use of scaling laws \citep[\eg][]{kaplan2020scaling, hoffmann2022chinchilla} suggests development is framed as an efficient resource allocation problem.
Evaluation does play a small role here: scaling laws relate resources (x-axis) with evaluated model performance (y-axis).
But the evaluation scope is narrow: scaling laws generally center accuracy for a single task (generally upstream language model perplexity), with \textit{predictability} of this relationship being the principal concern \citep[][\cf \citet{wei2022emergent}]{ganguli2022predictability}. 

In contrast, we argue evaluation does not currently exert similar overarching influence for language model development. 
While influential language models generally come from resource-rich institutions, they generally do not report evaluations on the same benchmarks.
\citet{liang2022helm} make this precise, showing that 30 prominent language models only evaluate on 17.9\% of the scenarios they introduce in HELM.
In fact, even for datasets that are most widely evaluated on across language modeling works at the time, \citet{liang2022helm} find that RTE is the unique dataset evaluated for in more than 50\% of the 32 works they consider that introduce language models (\eg GPT-3, GPT-NeoX, BLOOM, PaLM, Gopher, OPT, GLM).\footnote{This is likely a direct side effect of RTE being the unique dataset in both GLUE and SuperGLUE \citep{wang2019superglue, wang2019glue}.}
Given this status quo (\ie language models not even being evaluated on the same benchmarks), evaluation currently fails to achieve the widespread adoption to drive change, though recent high-profile evaluation efforts (\eg BIG-bench, the HuggingFace Evaluate library, HELM) may change this. 

\paragraph{Contrasting properties.}
Which forces orient NLP research is material: different forces profile differently.
Resources are distributed very unevenly, so resources orienting progress implies a narrow and particular subset of the community expresses out-sized influence in shaping the field's trajectory.
Further, by the nature of how these resource disparities came to be, these resource-rich actors tend to have specific incentives (\eg commercial interest) and demographics (\eg poor diversity), potentially causing them to advance more conservative agendas. 
In contrast, we believe evaluation structurally is better equipped to enable broader participation (\eg BIG-bench) and, critically, pluralism.
Different values can be simultaneously foregrounded in evaluations (\eg HELM \citep{liang2022helm} highlights values/desiderata such as accuracy, robustness, fairness, uncertainty, and efficiency).
For example, insofar as scaling laws drive language model development, greater pluralism would be achieved if scaling laws were studied, fit, and applied for a broader array of evaluation targets than just upstream accuracy/perplexity. 
\section{Legitimacy}
\label{sec:legitimacy}

Since evaluation accrues power, is this power legitimate?
And who does this power distribute to?

\paragraph{Legitimacy.}
As \citet{liang2022helm} note, evaluations are generally built by a small number of researchers, but then potentially go on to orient work in the broader research community.
Consequently, in arguing for the greater use of evaluation as a means for shifting power, we should question whether this implicitly recommends \textit{value imposition}: imposing values of the few onto the many.
However, we further recognize that the power of an evaluation is not principally determined by the evaluation creators, but is instead more directly contingent on the evaluation's adoption.
Consequently, for this power to emerge requires the consensual action of the early adopters, who choose to use the evaluation.
To an extent, this (voluntary) choice suggests that the power of evaluation is generally and, at least, initially legitimate.\footnote{As a matter of scope, we do not broach whether other forces in NLP research are similarly legitimate.} 

If the power of evaluation is legitimate, then what does this imply when evaluations are shown to have issues with respect to their validity, reliability, relevance, or appropriateness \citep[][\textit{inter alia}]
{
gururangan2018annotation,
kaushik2018rc,
ethayarajh2020classifier,
blodgett2021stereotyping,
aribandi2021reliable,
birhane2021pyrrhic}?
Here, we recognize that while the initial adoption of an evaluation is in most cases clearly legitimate, the subsequent sustained adoption can be more complicated.
In particular, we emphasize that evaluations tend to exhibit strong inertia: once an evaluation is widely adopted, it is hard for the evaluation to lose this adoption or for other evaluations to eclipse it \citep[\eg due to reviewing norms; ][]{dehghani2021lottery}, even when there are strong reasons to demote or deprecate the evaluation \citep{peng2021mitigating}. 
Most directly, we point to the strong norms of comparison in NLP, whereby model developers are expected to compare their models to prior models in head-to-head comparisons.
While generally a useful norm, this does promote a certain conservatism, especially when prior models (\ie those that are to be compared to) are not public or laborious to re-evaluate on new datasets, as new models can most easily be compared to old models on the evaluations used in prior work.
In this regard, paradigms where evaluations are continuously updated and refreshed (\eg the evaluation rounds in ANLI \citep{nie2020anli} and versions in HELM \citep{liang2022helm}; inherently dynamic evaluations like DynaBench \citep{kiela2021dynabench}) may more directly ensure the sustained power of specific evaluations is legitimate.

\paragraph{Distribution of power.}
Even if an evaluation's power is acquired legitimately, we should further question how the power distributes over different members of the community, especially as other forces (especially resources) are inequitably distributed.
\citet{koch2021reduced} show the distribution of evaluation developers is also uneven, aligning strongly with institutional privilege (\eg elite academic institutions like Stanford and Princeton, massive commercial organizations like Microsoft and Google). 
In part, this is likely a byproduct of the fact that evaluations themselves can be quite resource-intensive, especially when this scale is a virtue: 
ImageNet \citep{deng2009imagenet}, especially for its time, was exceedingly costly in both money and time; large-scale model evaluation on HELM costs almost $40$k USD in addition to $20$k A100 GPU hours \citep{liang2022helm}.

With that said, we have significant optimism that evaluation can realize more pluralistic visions.
Specifically, 
(i) the rise of foundation models in NLP has shifted the field towards few-shot evaluations \citep{brown2020gpt3, bragg2021flex}, which means evaluations need not include large-scale training subsets which constituted much of the cost for evaluations historically (\eg 80\%, or 80k+, of the examples in SQuAD \citep{rajpurkar2016squad} were allocated for training).
This suggests that their development should be more broadly accessible \citep[][\S4.4]{bommasani2021opportunities}, though the dynamics of their adoption are less clear.
Further, (ii) the practice of community-driven evaluation design has been successfully implemented in several instances: the EleutherAI LM Harness \cite{gao2021harness}, GEM \citep{gehrmann2021gem}, GEMv2 \citep{gehrmann2022gemv2}, BIG-Bench \citep{srivastava2022bigbench}, the Hugging Face Evaluate library \citep{werra2022evaluate}, with examples like Universal Dependencies \citep[UD; ][]{nivre2016universal, demarneffe2021universal} even pre-dating them for many years. 
In most cases, these efforts did not push a very clear directional change/agenda in research priorities (UD as a partial exception), but we believe future efforts could more explicitly exert power while learning from these prior efforts.
Finally, (iii) the community has grown to more properly recognize and value evaluation-type contributions (\eg~the NeurIPS datasets and benchmarks track, \cf~\citet{rogers2020resource}). 
That is, while we argue evaluation's power is currently waning relative to resources, suggesting a trend towards less pluralism, we simultaneously believe the conditions are ripe for renewed commitment to evaluation to reverse this trajectory.
\section{Conclusion}
\label{sec:conclusion}

Evaluation wields power: we believe the community is largely aware of this, yet we foreground this power to understand how evaluation drives change. 
This perspective leads us to three conclusions:
(i) adoption imbues evaluation with its power,
(ii) evaluation's power relative to other competing social forces appears to be diminishing,
and yet (iii) evaluation has attractive qualities, especially under current conditions, as a force for change relative to other forces with growing power.

Overall, we hope the community reflects on the mantra "evaluation for change" and the Gandhian maxim "build evaluations for the change you wish to see in the world".\footnote{Remark: The associated quote that this riffs off of was never actually said by Mahatma Gandhi, in spite of this popular misconception. But we believe its essence aligns well with Gandhi's character nonetheless.}

\section*{Limitations}
This work puts forth a position: by the nature of a position paper, the work is deliberately intended to be evocative and opinionated, in some places not having unequivocal evidence for certain claims. 
This presents a clear limitation: the analysis presented may diverge from the realities of NLP at present or in the future, namely if the assumptions/conditions presented themselves prove to be untrue in practice.
Nonetheless, we believe centering power and change, and understanding evaluation as a political and sociological phenomenon, is likely to be useful under all conditions. 

Further, in understanding the qualities of evaluation relative to other social forces, we directly suggest that evaluation is more readily operationalized in more pluralistic ways than other key forces (primarily resources).
While initial efforts indicate the potential for such holistic approaches that reflect many different desiderata \citep{liang2022helm} as well as participatory approaches that permit contribution from different entities \citep[\eg][]{srivastava2022bigbench}, it is still unclear how much adoption such approaches will get, and therefore how much power they will acquire.
That is, the extent to which evaluation can realize this pluralistic vision still largely remains an unresolved aspiration than a readily realizable certainty.
And, conversely, we do note that while current practices potentially put pluralism and resources at odds, they may be mutually compatible in other regimes (\eg decentralized training through the pooling of shared/volunteered compute \citep{yuan2022decentralized}, open-source software development \citep{wolf2020transformers, gao2021harness, werra2022evaluate}).

Finally, we do not discuss other forces that we believe have not exhibited strong influence on NLP research thus far, in favor of allocating focus to evaluation and resources, which have had clear influence.
To enumerate some of these other (potential) forces, we specifically note 
(i) research norms, 
(ii) policy and regulation,
and (iii) auditing/advocacy.
For (i), we note that while the NLP research community has many established norms (\eg reproducibility checklists, peer review guidelines, conference organization structure, policies on respectful conduct), most of these do not directly/significantly influence what research topics different researchers work on.
We do note that is possible in the future that certain norms \citep[\eg the access to training data or model checkpoints;][]{liang2022norms} would influence what research is conducted (\eg we may have not seen as much work on the learning dynamics of language models and/or memorization of training data due to the relative inaccessibility of intermediary checkpoints and training data until recently). 
For (ii), we note that policy and regulatory efforts have had little to no salient impact on the deployment of most language technologies, let alone NLP research, to our knowledge.
With that said, much as efforts like GDPR and privacy legislation has impacted scientific research on privacy \citep[\eg work that operationalizes the right to be forgotten as in][]{ginart2019right}, similar trends could occur in NLP research (\eg in response to the EU AI Act).
Akin to (ii), for (iii), we also have seen fairly little impact from auditing/advocacy work on NLP research to our knowledge.
But, much as work on auditing/advocacy around face recognition \citep[][\textit{inter alia}]{buolamwini2018gender, raji2019actionable, raji2020saving} influenced research in the computer vision community, we could see similar trends in NLP (\eg in response to auditing/advocacy intervention around language models). 

\section*{Ethics Statement}
We do not find serious risks or ethical concerns with this work.
We do note this work advances a specific position, which we clearly identify. 
It should not be assumed there is consensus in the community (or beyond) on any account for evaluation, let alone the account on power that we espouse. 
In this regard, we actively solicit response and interrogation of the positions presented in this work, especially given myriad relevant analyses of evaluation/measurement/benchmarking exist in other parts of AI, computer science, linguistics, and other disciplines.  

\section*{Acknowledgements}
I would like to thank 
Alex Hanna,
Ben Recht,
Chris Manning,
Chris Potts,
Claire Cardie,
Dan Jurafsky,
Deb Raji,
Emily Denton,
Henrik Kugelberg,
Jacob Andreas,
Jacob Steinhardt,
John Hewitt,
Judy Shen,
Kawin Ethayarajh,
Nelson Liu,
Rediet Abebe,
Rohan Taori,
Sam Bowman,
Stella Biderman,
Tal Linzen,
Yann Dubois,
and
Yoav Goldberg
for being specific inspirations, whose writings and thoughts helped develop my current position on evaluation, with special thanks to Percy Liang.
I would like to thank the CRFM community; 
the experience of designing and building HELM \citep{liang2022helm} in particular helped sharpen my belief in this philosophy towards evaluation.
I am  supported by the NSF Graduate Research Fellowship Program under grant number DGE-1655618.

\bibliography{anthology,custom}

\begin{thebibliography}{69}
\expandafter\ifx\csname natexlab\endcsname\relax\def\natexlab#1{#1}\fi

\bibitem[{Abebe et~al.(2020)Abebe, Barocas, Kleinberg, Levy, Raghavan, and
  Robinson}]{abebe2020roles}
Rediet Abebe, Solon Barocas, Jon Kleinberg, Karen Levy, Manish Raghavan, and
  David~G. Robinson. 2020.
\newblock \href {https://doi.org/10.1145/3351095.3372871} {Roles for computing
  in social change}.
\newblock In \emph{Proceedings of the 2020 Conference on Fairness,
  Accountability, and Transparency}, FAT* '20, page 252–260, New York, NY,
  USA. Association for Computing Machinery.

\bibitem[{Aribandi et~al.(2021)Aribandi, Tay, and
  Metzler}]{aribandi2021reliable}
Vamsi Aribandi, Yi~Tay, and Donald Metzler. 2021.
\newblock \href {https://doi.org/10.18653/v1/2021.findings-acl.155} {How
  reliable are model diagnostics?}
\newblock In \emph{Findings of the Association for Computational Linguistics:
  ACL-IJCNLP 2021}, pages 1778--1785, Online. Association for Computational
  Linguistics.

\bibitem[{Birhane et~al.(2022)Birhane, Kalluri, Card, Agnew, Dotan, and
  Bao}]{birhane2022values}
Abeba Birhane, Pratyusha Kalluri, Dallas Card, William Agnew, Ravit Dotan, and
  Michelle Bao. 2022.
\newblock \href {https://doi.org/10.1145/3531146.3533083} {The values encoded
  in machine learning research}.
\newblock In \emph{2022 ACM Conference on Fairness, Accountability, and
  Transparency}, FAccT '22, page 173–184, New York, NY, USA. Association for
  Computing Machinery.

\bibitem[{Birhane and Prabhu(2021)}]{birhane2021pyrrhic}
Abeba Birhane and Vinay~Uday Prabhu. 2021.
\newblock \href {https://doi.org/10.1109/WACV48630.2021.00158} {Large image
  datasets: A pyrrhic win for computer vision?}
\newblock In \emph{2021 IEEE Winter Conference on Applications of Computer
  Vision (WACV)}, pages 1536--1546.

\bibitem[{Blodgett et~al.(2021)Blodgett, Lopez, Olteanu, Sim, and
  Wallach}]{blodgett2021stereotyping}
Su~Lin Blodgett, Gilsinia Lopez, Alexandra Olteanu, Robert Sim, and Hanna
  Wallach. 2021.
\newblock \href {https://doi.org/10.18653/v1/2021.acl-long.81} {Stereotyping
  {N}orwegian salmon: An inventory of pitfalls in fairness benchmark datasets}.
\newblock In \emph{Proceedings of the 59th Annual Meeting of the Association
  for Computational Linguistics and the 11th International Joint Conference on
  Natural Language Processing (Volume 1: Long Papers)}, pages 1004--1015,
  Online. Association for Computational Linguistics.

\bibitem[{Bommasani et~al.(2021)Bommasani, Hudson, Adeli, Altman, Arora, von
  Arx, Bernstein, Bohg, Bosselut, Brunskill, Brynjolfsson, Buch, Card,
  Castellon, Chatterji, Chen, Creel, Davis, Demszky, Donahue, Doumbouya,
  Durmus, Ermon, Etchemendy, Ethayarajh, Fei-Fei, Finn, Gale, Gillespie, Goel,
  Goodman, Grossman, Guha, Hashimoto, Henderson, Hewitt, Ho, Hong, Hsu, Huang,
  Icard, Jain, Jurafsky, Kalluri, Karamcheti, Keeling, Khani, Khattab, Koh,
  Krass, Krishna, Kuditipudi, Kumar, Ladhak, Lee, Lee, Leskovec, Levent, Li,
  Li, Ma, Malik, Manning, Mirchandani, Mitchell, Munyikwa, Nair, Narayan,
  Narayanan, Newman, Nie, Niebles, Nilforoshan, Nyarko, Ogut, Orr,
  Papadimitriou, Park, Piech, Portelance, Potts, Raghunathan, Reich, Ren, Rong,
  Roohani, Ruiz, Ryan, R{\`e}, Sadigh, Sagawa, Santhanam, Shih, Srinivasan,
  Tamkin, Taori, Thomas, Tram{\`e}r, Wang, Wang, Wu, Wu, Wu, Xie, Yasunaga,
  You, Zaharia, Zhang, Zhang, Zhang, Zhang, Zheng, Zhou, and
  Liang}]{bommasani2021opportunities}
Rishi Bommasani, Drew~A. Hudson, Ehsan Adeli, Russ Altman, Simran Arora, Sydney
  von Arx, Michael~S. Bernstein, Jeannette Bohg, Antoine Bosselut, Emma
  Brunskill, Erik Brynjolfsson, S.~Buch, Dallas Card, Rodrigo Castellon,
  Niladri~S. Chatterji, Annie~S. Chen, Kathleen~A. Creel, Jared Davis, Dora
  Demszky, Chris Donahue, Moussa Doumbouya, Esin Durmus, Stefano Ermon, John
  Etchemendy, Kawin Ethayarajh, Li~Fei-Fei, Chelsea Finn, Trevor Gale,
  Lauren~E. Gillespie, Karan Goel, Noah~D. Goodman, Shelby Grossman, Neel Guha,
  Tatsunori Hashimoto, Peter Henderson, John Hewitt, Daniel~E. Ho, Jenny Hong,
  Kyle Hsu, Jing Huang, Thomas~F. Icard, Saahil Jain, Dan Jurafsky, Pratyusha
  Kalluri, Siddharth Karamcheti, Geoff Keeling, Fereshte Khani, O.~Khattab,
  Pang~Wei Koh, Mark~S. Krass, Ranjay Krishna, Rohith Kuditipudi, Ananya Kumar,
  Faisal Ladhak, Mina Lee, Tony Lee, Jure Leskovec, Isabelle Levent, Xiang~Lisa
  Li, Xuechen Li, Tengyu Ma, Ali Malik, Christopher~D. Manning, Suvir~P.
  Mirchandani, Eric Mitchell, Zanele Munyikwa, Suraj Nair, Avanika Narayan,
  Deepak Narayanan, Benjamin Newman, Allen Nie, Juan~Carlos Niebles, Hamed
  Nilforoshan, Julian~F. Nyarko, Giray Ogut, Laurel Orr, Isabel Papadimitriou,
  Joon~Sung Park, Chris Piech, Eva Portelance, Christopher Potts, Aditi
  Raghunathan, Robert Reich, Hongyu Ren, Frieda Rong, Yusuf~H. Roohani, Camilo
  Ruiz, Jack Ryan, Christopher R{\`e}, Dorsa Sadigh, Shiori Sagawa, Keshav
  Santhanam, Andy Shih, Krishna~Parasuram Srinivasan, Alex Tamkin, Rohan Taori,
  Armin~W. Thomas, Florian Tram{\`e}r, Rose~E. Wang, William Wang, Bohan Wu,
  Jiajun Wu, Yuhuai Wu, Sang~Michael Xie, Michihiro Yasunaga, Jiaxuan You,
  Matei~A. Zaharia, Michael Zhang, Tianyi Zhang, Xikun Zhang, Yuhui Zhang,
  Lucia Zheng, Kaitlyn Zhou, and Percy Liang. 2021.
\newblock \href {https://crfm.stanford.edu/assets/report.pdf} {On the
  opportunities and risks of foundation models}.
\newblock \emph{ArXiv}.

\bibitem[{Bowman and Dahl(2021)}]{bowman2021fix}
Samuel~R. Bowman and George Dahl. 2021.
\newblock \href {https://doi.org/10.18653/v1/2021.naacl-main.385} {What will it
  take to fix benchmarking in natural language understanding?}
\newblock In \emph{Proceedings of the 2021 Conference of the North American
  Chapter of the Association for Computational Linguistics: Human Language
  Technologies}, pages 4843--4855, Online. Association for Computational
  Linguistics.

\bibitem[{Bragg et~al.(2021)Bragg, Cohan, Lo, and Beltagy}]{bragg2021flex}
Jonathan Bragg, Arman Cohan, Kyle Lo, and Iz~Beltagy. 2021.
\newblock \href {https://openreview.net/forum?id=_WnGcwXLYOE} {{FLEX}: Unifying
  evaluation for few-shot {NLP}}.
\newblock In \emph{Advances in Neural Information Processing Systems}.

\bibitem[{Brown et~al.(2020)Brown, Mann, Ryder, Subbiah, Kaplan, Dhariwal,
  Neelakantan, Shyam, Sastry, Askell, Agarwal, Herbert-Voss, Krueger, Henighan,
  Child, Ramesh, Ziegler, Wu, Winter, Hesse, Chen, Sigler, Litwin, Gray, Chess,
  Clark, Berner, McCandlish, Radford, Sutskever, and Amodei}]{brown2020gpt3}
Tom Brown, Benjamin Mann, Nick Ryder, Melanie Subbiah, Jared~D Kaplan, Prafulla
  Dhariwal, Arvind Neelakantan, Pranav Shyam, Girish Sastry, Amanda Askell,
  Sandhini Agarwal, Ariel Herbert-Voss, Gretchen Krueger, Tom Henighan, Rewon
  Child, Aditya Ramesh, Daniel Ziegler, Jeffrey Wu, Clemens Winter, Chris
  Hesse, Mark Chen, Eric Sigler, Mateusz Litwin, Scott Gray, Benjamin Chess,
  Jack Clark, Christopher Berner, Sam McCandlish, Alec Radford, Ilya Sutskever,
  and Dario Amodei. 2020.
\newblock \href
  {https://proceedings.neurips.cc/paper/2020/file/1457c0d6bfcb4967418bfb8ac142f64a-Paper.pdf}
  {Language models are few-shot learners}.
\newblock In \emph{Advances in Neural Information Processing Systems},
  volume~33, pages 1877--1901. Curran Associates, Inc.

\bibitem[{Buolamwini and Gebru(2018)}]{buolamwini2018gender}
Joy Buolamwini and Timnit Gebru. 2018.
\newblock \href {https://proceedings.mlr.press/v81/buolamwini18a.html} {Gender
  shades: Intersectional accuracy disparities in commercial gender
  classification}.
\newblock In \emph{Proceedings of the 1st Conference on Fairness,
  Accountability and Transparency}, volume~81 of \emph{Proceedings of Machine
  Learning Research}, pages 77--91. PMLR.

\bibitem[{de~Marneffe et~al.(2021)de~Marneffe, Manning, Nivre, and
  Zeman}]{demarneffe2021universal}
Marie-Catherine de~Marneffe, Christopher~D. Manning, Joakim Nivre, and Daniel
  Zeman. 2021.
\newblock \href {https://doi.org/10.1162/coli_a_00402} {{U}niversal
  {D}ependencies}.
\newblock \emph{Computational Linguistics}, 47(2):255--308.

\bibitem[{Dehghani et~al.(2021)Dehghani, Tay, Gritsenko, Zhao, Houlsby, Diaz,
  Metzler, and Vinyals}]{dehghani2021lottery}
Mostafa Dehghani, Yi~Tay, Alexey~A. Gritsenko, Zhe Zhao, Neil Houlsby, Fernando
  Diaz, Donald Metzler, and Oriol Vinyals. 2021.
\newblock The benchmark lottery.
\newblock \emph{ArXiv}, abs/2107.07002.

\bibitem[{Deng et~al.(2009)Deng, Dong, Socher, Li, Li, and
  Fei-Fei}]{deng2009imagenet}
Jia Deng, Wei Dong, Richard Socher, Li-Jia Li, Kai Li, and Li~Fei-Fei. 2009.
\newblock \href {https://doi.org/10.1109/CVPR.2009.5206848} {Imagenet: A
  large-scale hierarchical image database}.
\newblock In \emph{2009 IEEE Conference on Computer Vision and Pattern
  Recognition}, pages 248--255.

\bibitem[{Denton et~al.(2021)Denton, Hanna, Amironesei, Smart, and
  Nicole}]{denton2021genealogy}
Emily Denton, Alex Hanna, Razvan Amironesei, Andrew Smart, and Hilary Nicole.
  2021.
\newblock \href {https://doi.org/10.1177/20539517211035955} {On the genealogy
  of machine learning datasets: A critical history of imagenet}.
\newblock \emph{Big Data \& Society}, 8(2):20539517211035955.

\bibitem[{Dotan and Milli(2020)}]{dotan2020value}
Ravit Dotan and Smitha Milli. 2020.
\newblock Value-laden disciplinary shifts in machine learning.
\newblock \emph{Proceedings of the 2020 Conference on Fairness, Accountability,
  and Transparency}.

\bibitem[{Ethayarajh(2020)}]{ethayarajh2020classifier}
Kawin Ethayarajh. 2020.
\newblock \href {https://doi.org/10.18653/v1/2020.acl-main.262} {Is your
  classifier actually biased? measuring fairness under uncertainty with
  bernstein bounds}.
\newblock In \emph{Proceedings of the 58th Annual Meeting of the Association
  for Computational Linguistics}, pages 2914--2919, Online. Association for
  Computational Linguistics.

\bibitem[{Ethayarajh and Jurafsky(2020)}]{ethayarajh2020utility}
Kawin Ethayarajh and Dan Jurafsky. 2020.
\newblock \href {https://doi.org/10.18653/v1/2020.emnlp-main.393} {Utility is
  in the eye of the user: A critique of {NLP} leaderboards}.
\newblock In \emph{Proceedings of the 2020 Conference on Empirical Methods in
  Natural Language Processing (EMNLP)}, pages 4846--4853, Online. Association
  for Computational Linguistics.

\bibitem[{Ganguli et~al.(2022)Ganguli, Hernandez, Lovitt, Askell, Bai, Chen,
  Conerly, Dassarma, Drain, Elhage, El~Showk, Fort, Hatfield-Dodds, Henighan,
  Johnston, Jones, Joseph, Kernian, Kravec, Mann, Nanda, Ndousse, Olsson,
  Amodei, Brown, Kaplan, McCandlish, Olah, Amodei, and
  Clark}]{ganguli2022predictability}
Deep Ganguli, Danny Hernandez, Liane Lovitt, Amanda Askell, Yuntao Bai, Anna
  Chen, Tom Conerly, Nova Dassarma, Dawn Drain, Nelson Elhage, Sheer El~Showk,
  Stanislav Fort, Zac Hatfield-Dodds, Tom Henighan, Scott Johnston, Andy Jones,
  Nicholas Joseph, Jackson Kernian, Shauna Kravec, Ben Mann, Neel Nanda, Kamal
  Ndousse, Catherine Olsson, Daniela Amodei, Tom Brown, Jared Kaplan, Sam
  McCandlish, Christopher Olah, Dario Amodei, and Jack Clark. 2022.
\newblock \href {https://doi.org/10.1145/3531146.3533229} {Predictability and
  surprise in large generative models}.
\newblock In \emph{2022 ACM Conference on Fairness, Accountability, and
  Transparency}, FAccT '22, page 1747–1764, New York, NY, USA. Association
  for Computing Machinery.

\bibitem[{Gao et~al.(2021)Gao, Tow, Biderman, Black, DiPofi, Foster, Golding,
  Hsu, McDonell, Muennighoff, Phang, Reynolds, Tang, Thite, Wang, Wang, and
  Zou}]{gao2021harness}
Leo Gao, Jonathan Tow, Stella Biderman, Sid Black, Anthony DiPofi, Charles
  Foster, Laurence Golding, Jeffrey Hsu, Kyle McDonell, Niklas Muennighoff,
  Jason Phang, Laria Reynolds, Eric Tang, Anish Thite, Ben Wang, Kevin Wang,
  and Andy Zou. 2021.
\newblock A framework for few-shot language model evaluation.
\newblock \emph{Version v0. 0.1. Sept}.

\bibitem[{Gehrmann et~al.(2021)Gehrmann, Adewumi, Aggarwal, Ammanamanchi,
  Aremu, Bosselut, Chandu, Clinciu, Das, Dhole, Du, Durmus, Du{\v{s}}ek,
  Emezue, Gangal, Garbacea, Hashimoto, Hou, Jernite, Jhamtani, Ji, Jolly, Kale,
  Kumar, Ladhak, Madaan, Maddela, Mahajan, Mahamood, Majumder, Martins,
  McMillan-Major, Mille, van Miltenburg, Nadeem, Narayan, Nikolaev,
  Niyongabo~Rubungo, Osei, Parikh, Perez-Beltrachini, Rao, Raunak, Rodriguez,
  Santhanam, Sedoc, Sellam, Shaikh, Shimorina, Sobrevilla~Cabezudo, Strobelt,
  Subramani, Xu, Yang, Yerukola, and Zhou}]{gehrmann2021gem}
Sebastian Gehrmann, Tosin Adewumi, Karmanya Aggarwal, Pawan~Sasanka
  Ammanamanchi, Anuoluwapo Aremu, Antoine Bosselut, Khyathi~Raghavi Chandu,
  Miruna-Adriana Clinciu, Dipanjan Das, Kaustubh Dhole, Wanyu Du, Esin Durmus,
  Ond{\v{r}}ej Du{\v{s}}ek, Chris~Chinenye Emezue, Varun Gangal, Cristina
  Garbacea, Tatsunori Hashimoto, Yufang Hou, Yacine Jernite, Harsh Jhamtani,
  Yangfeng Ji, Shailza Jolly, Mihir Kale, Dhruv Kumar, Faisal Ladhak, Aman
  Madaan, Mounica Maddela, Khyati Mahajan, Saad Mahamood, Bodhisattwa~Prasad
  Majumder, Pedro~Henrique Martins, Angelina McMillan-Major, Simon Mille, Emiel
  van Miltenburg, Moin Nadeem, Shashi Narayan, Vitaly Nikolaev, Andre
  Niyongabo~Rubungo, Salomey Osei, Ankur Parikh, Laura Perez-Beltrachini,
  Niranjan~Ramesh Rao, Vikas Raunak, Juan~Diego Rodriguez, Sashank Santhanam,
  Jo{\~a}o Sedoc, Thibault Sellam, Samira Shaikh, Anastasia Shimorina,
  Marco~Antonio Sobrevilla~Cabezudo, Hendrik Strobelt, Nishant Subramani, Wei
  Xu, Diyi Yang, Akhila Yerukola, and Jiawei Zhou. 2021.
\newblock \href {https://doi.org/10.18653/v1/2021.gem-1.10} {The {GEM}
  benchmark: Natural language generation, its evaluation and metrics}.
\newblock In \emph{Proceedings of the 1st Workshop on Natural Language
  Generation, Evaluation, and Metrics (GEM 2021)}, pages 96--120, Online.
  Association for Computational Linguistics.

\bibitem[{Gehrmann et~al.(2022)Gehrmann, Bhattacharjee, Mahendiran, Wang,
  Papangelis, Madaan, McMillan-Major, Shvets, Upadhyay, Yao, Wilie,
  Bhagavatula, You, Thomson, Garbacea, Wang, Deutsch, Xiong, Jin, Gkatzia,
  Radev, Clark, Durmus, Ladhak, Ginter, Winata, Strobelt, Hayashi, Novikova,
  Kanerva, Chim, Zhou, Clive, Maynez, Sedoc, Juraska, Dhole, Chandu, Ribeiro,
  Tunstall, Zhang, Pushkarna, Creutz, White, Kale, Eddine, Daheim, Subramani,
  Dusek, Liang, Ammanamanchi, Zhu, Puduppully, Kriz, Shahriyar, Cardenas,
  Mahamood, Osei, Cahyawijaya, vStajner, Montella, Shailza, Jolly, Mille,
  Hasan, Shen, Adewumi, Raunak, Raheja, Nikolaev, Tsai, Jernite, Xu, Sang, Liu,
  and Hou}]{gehrmann2022gemv2}
Sebastian Gehrmann, Abhik Bhattacharjee, Abinaya Mahendiran, Alex Wang,
  Alexandros Papangelis, Aman Madaan, Angelina McMillan-Major, Anna~V. Shvets,
  Ashish Upadhyay, Bingsheng Yao, Bryan Wilie, Chandra Bhagavatula, Chaobin
  You, Craig Thomson, Cristina Garbacea, Dakuo Wang, Daniel Deutsch, Deyi
  Xiong, Di~Jin, Dimitra Gkatzia, Dragomir Radev, Elizabeth Clark, Esin Durmus,
  Faisal Ladhak, Filip Ginter, Genta~Indra Winata, Hendrik Strobelt, Hiroaki
  Hayashi, Jekaterina Novikova, Jenna Kanerva, Jenny Chim, Jiawei Zhou, Jordan
  Clive, Joshua Maynez, Jo{\~a}o Sedoc, Juraj Juraska, Kaustubh~D. Dhole,
  Khyathi~Raghavi Chandu, Leonardo F.~R. Ribeiro, Lewis Tunstall, Li~Zhang,
  Mahima Pushkarna, Mathias Creutz, Michael White, Mihir Kale, Moussa~Kamal
  Eddine, Nico Daheim, Nishant Subramani, Ondrej Dusek, Paul~Pu Liang,
  Pawan~Sasanka Ammanamanchi, Qinqin Zhu, Ratish Puduppully, Reno Kriz, Rifat
  Shahriyar, Ronald Cardenas, Saad Mahamood, Salomey Osei, Samuel Cahyawijaya,
  Sanja vStajner, S{\'e}bastien Montella, Shailza, Shailza Jolly, Simon Mille,
  Tahmid Hasan, Tianhao Shen, Tosin~P. Adewumi, Vikas Raunak, Vipul Raheja,
  Vitaly Nikolaev, Vivian Tsai, Yacine Jernite, Yi~Xu, Yisi Sang, Yixin Liu,
  and Yufang Hou. 2022.
\newblock Gemv2: Multilingual nlg benchmarking in a single line of code.
\newblock \emph{ArXiv}, abs/2206.11249.

\bibitem[{Ginart et~al.(2019)Ginart, Guan, Valiant, and Zou}]{ginart2019right}
Antonio Ginart, Melody Guan, Gregory Valiant, and James~Y Zou. 2019.
\newblock \href
  {https://proceedings.neurips.cc/paper/2019/file/cb79f8fa58b91d3af6c9c991f63962d3-Paper.pdf}
  {Making ai forget you: Data deletion in machine learning}.
\newblock In \emph{Advances in Neural Information Processing Systems},
  volume~32. Curran Associates, Inc.

\bibitem[{Grosz et~al.(1983)Grosz, Joshi, and Weinstein}]{grosz1983providing}
Barbara~J. Grosz, Aravind~K. Joshi, and Scott Weinstein. 1983.
\newblock \href {https://doi.org/10.3115/981311.981320} {{Providing a Unified
  Account of Definite Noun Phrases in Discourse}}.
\newblock In \emph{21st Annual Meeting of the Association for Computational
  Linguistics}, pages 44--50, Cambridge, Massachusetts, USA. Association for
  Computational Linguistics.

\bibitem[{Gururangan et~al.(2018)Gururangan, Swayamdipta, Levy, Schwartz,
  Bowman, and Smith}]{gururangan2018annotation}
Suchin Gururangan, Swabha Swayamdipta, Omer Levy, Roy Schwartz, Samuel Bowman,
  and Noah~A. Smith. 2018.
\newblock \href {https://doi.org/10.18653/v1/N18-2017} {Annotation artifacts in
  natural language inference data}.
\newblock In \emph{Proceedings of the 2018 Conference of the North {A}merican
  Chapter of the Association for Computational Linguistics: Human Language
  Technologies, Volume 2 (Short Papers)}, pages 107--112, New Orleans,
  Louisiana. Association for Computational Linguistics.

\bibitem[{Hallinan and Striphas(2016)}]{hallinan2016netflix}
Blake Hallinan and Ted Striphas. 2016.
\newblock \href {https://doi.org/10.1177/1461444814538646} {Recommended for
  you: The netflix prize and the production of algorithmic culture}.
\newblock \emph{New Media \& Society}, 18(1):117--137.

\bibitem[{Hoffmann et~al.(2022)Hoffmann, Borgeaud, Mensch, Buchatskaya, Cai,
  Rutherford, de~las Casas, Hendricks, Welbl, Clark, Hennigan, Noland,
  Millican, van~den Driessche, Damoc, Guy, Osindero, Simonyan, Elsen, Vinyals,
  Rae, and Sifre}]{hoffmann2022chinchilla}
Jordan Hoffmann, Sebastian Borgeaud, Arthur Mensch, Elena Buchatskaya, Trevor
  Cai, Eliza Rutherford, Diego de~las Casas, Lisa~Anne Hendricks, Johannes
  Welbl, Aidan Clark, Tom Hennigan, Eric Noland, Katherine Millican, George
  van~den Driessche, Bogdan Damoc, Aurelia Guy, Simon Osindero, Karen Simonyan,
  Erich Elsen, Oriol Vinyals, Jack~William Rae, and Laurent Sifre. 2022.
\newblock \href {https://openreview.net/forum?id=iBBcRUlOAPR} {An empirical
  analysis of compute-optimal large language model training}.
\newblock In \emph{Advances in Neural Information Processing Systems}.

\bibitem[{Hutchinson et~al.(2021)Hutchinson, Smart, Hanna, Denton, Greer,
  Kjartansson, Barnes, and Mitchell}]{hutchinson2021towards}
Ben Hutchinson, Andrew Smart, Alex Hanna, Emily Denton, Christina Greer, Oddur
  Kjartansson, Parker Barnes, and Margaret Mitchell. 2021.
\newblock \href {https://doi.org/10.1145/3442188.3445918} {Towards
  accountability for machine learning datasets: Practices from software
  engineering and infrastructure}.
\newblock In \emph{Proceedings of the 2021 ACM Conference on Fairness,
  Accountability, and Transparency}, FAccT '21, page 560–575, New York, NY,
  USA. Association for Computing Machinery.

\bibitem[{Jacobs and Wallach(2021)}]{jacobs2021measurement}
Abigail~Z. Jacobs and Hanna Wallach. 2021.
\newblock \href {https://doi.org/10.1145/3442188.3445901} {Measurement and
  fairness}.
\newblock In \emph{Proceedings of the 2021 ACM Conference on Fairness,
  Accountability, and Transparency}, FAccT '21, page 375–385, New York, NY,
  USA. Association for Computing Machinery.

\bibitem[{Joshi(1969)}]{joshi1969sag}
Aravind~K. Joshi. 1969.
\newblock \href {https://aclanthology.org/C69-4701} {{Properties of Formal
  Grammars with Mixed Type of Rules and their Linguistic Relevance}}.
\newblock In \emph{{I}nternational {C}onference on {C}omputational
  {L}inguistics {COLING} 1969: Preprint No. 47}, S{\aa}nga S{\"a}by, Sweden.

\bibitem[{Joshi et~al.(1972)Joshi, Kosaraju, and Yamada}]{joshi1972sag}
Aravind~K. Joshi, S.~Rao Kosaraju, and H.M. Yamada. 1972.
\newblock \href {https://doi.org/https://doi.org/10.1016/S0019-9958(72)90051-4}
  {String adjunct grammars: I. local and distributed adjunction}.
\newblock \emph{Information and Control}, 21(2):93--116.

\bibitem[{Joshi and Schabes(1989)}]{joshi1989evaluation}
Aravind~K. Joshi and Yves Schabes. 1989.
\newblock \href {https://aclanthology.org/H89-2053} {An evaluation of
  lexicalization in parsing}.
\newblock In \emph{Speech and Natural Language: Proceedings of a Workshop Held
  at Cape Cod, Massachusetts, October 15-18, 1989}.

\bibitem[{Kaplan et~al.(2020)Kaplan, McCandlish, Henighan, Brown, Chess, Child,
  Gray, Radford, Wu, and Amodei}]{kaplan2020scaling}
Jared Kaplan, Sam McCandlish, T.~J. Henighan, Tom~B. Brown, Benjamin Chess,
  Rewon Child, Scott Gray, Alec Radford, Jeff Wu, and Dario Amodei. 2020.
\newblock Scaling laws for neural language models.
\newblock \emph{ArXiv}, abs/2001.08361.

\bibitem[{Kaushik and Lipton(2018)}]{kaushik2018rc}
Divyansh Kaushik and Zachary~C. Lipton. 2018.
\newblock \href {https://doi.org/10.18653/v1/D18-1546} {How much reading does
  reading comprehension require? a critical investigation of popular
  benchmarks}.
\newblock In \emph{Proceedings of the 2018 Conference on Empirical Methods in
  Natural Language Processing}, pages 5010--5015, Brussels, Belgium.
  Association for Computational Linguistics.

\bibitem[{Kiela et~al.(2021)Kiela, Bartolo, Nie, Kaushik, Geiger, Wu, Vidgen,
  Prasad, Singh, Ringshia, Ma, Thrush, Riedel, Waseem, Stenetorp, Jia, Bansal,
  Potts, and Williams}]{kiela2021dynabench}
Douwe Kiela, Max Bartolo, Yixin Nie, Divyansh Kaushik, Atticus Geiger,
  Zhengxuan Wu, Bertie Vidgen, Grusha Prasad, Amanpreet Singh, Pratik Ringshia,
  Zhiyi Ma, Tristan Thrush, Sebastian Riedel, Zeerak Waseem, Pontus Stenetorp,
  Robin Jia, Mohit Bansal, Christopher Potts, and Adina Williams. 2021.
\newblock \href {https://doi.org/10.18653/v1/2021.naacl-main.324} {Dynabench:
  Rethinking benchmarking in {NLP}}.
\newblock In \emph{Proceedings of the 2021 Conference of the North American
  Chapter of the Association for Computational Linguistics: Human Language
  Technologies}, pages 4110--4124, Online. Association for Computational
  Linguistics.

\bibitem[{Koch et~al.(2021)Koch, Denton, Hanna, and Foster}]{koch2021reduced}
Bernard Koch, Emily Denton, Alex Hanna, and Jacob~Gates Foster. 2021.
\newblock \href {https://openreview.net/forum?id=zNQBIBKJRkd} {Reduced, reused
  and recycled: The life of a dataset in machine learning research}.
\newblock In \emph{Thirty-fifth Conference on Neural Information Processing
  Systems Datasets and Benchmarks Track (Round 2)}.

\bibitem[{Krizhevsky et~al.(2012)Krizhevsky, Sutskever, and
  Hinton}]{krizhevsky2012alexnet}
Alex Krizhevsky, Ilya Sutskever, and Geoffrey~E Hinton. 2012.
\newblock \href
  {https://proceedings.neurips.cc/paper/2012/file/c399862d3b9d6b76c8436e924a68c45b-Paper.pdf}
  {Imagenet classification with deep convolutional neural networks}.
\newblock In \emph{Advances in Neural Information Processing Systems},
  volume~25. Curran Associates, Inc.

\bibitem[{Liang et~al.(2022{\natexlab{a}})Liang, Bommasani, Creel, and
  Reich}]{liang2022norms}
Percy Liang, Rishi Bommasani, Kathleen~A. Creel, and Rob Reich.
  2022{\natexlab{a}}.
\newblock \href {https://crfm.stanford.edu/2022/05/17/community-norms.html}
  {The time is now to develop community norms for the release of foundation
  models}.

\bibitem[{Liang et~al.(2022{\natexlab{b}})Liang, Bommasani, Lee, Tsipras,
  Soylu, Yasunaga, Zhang, Narayanan, Wu, Kumar, Newman, Yuan, Yan, Zhang,
  Cosgrove, Manning, R'e, Acosta-Navas, Hudson, Zelikman, Durmus, Ladhak, Rong,
  Ren, Yao, Wang, Santhanam, Orr, Zheng, Yuksekgonul, Suzgun, Kim, Guha,
  Chatterji, Khattab, Henderson, Huang, Chi, Xie, Santurkar, Ganguli,
  Hashimoto, Icard, Zhang, Chaudhary, Wang, Li, Mai, Zhang, and
  Koreeda}]{liang2022helm}
Percy Liang, Rishi Bommasani, Tony Lee, Dimitris Tsipras, Dilara Soylu,
  Michihiro Yasunaga, Yian Zhang, Deepak Narayanan, Yuhuai Wu, Ananya Kumar,
  Benjamin Newman, Binhang Yuan, Bobby Yan, Ce~Zhang, Christian Cosgrove,
  Christopher~D. Manning, Christopher R'e, Diana Acosta-Navas, Drew~A. Hudson,
  E.~Zelikman, Esin Durmus, Faisal Ladhak, Frieda Rong, Hongyu Ren, Huaxiu Yao,
  Jue Wang, Keshav Santhanam, Laurel~J. Orr, Lucia Zheng, Mert Yuksekgonul,
  Mirac Suzgun, Nathan~S. Kim, Neel Guha, Niladri~S. Chatterji, O.~Khattab,
  Peter Henderson, Qian Huang, Ryan Chi, Sang~Michael Xie, Shibani Santurkar,
  Surya Ganguli, Tatsunori Hashimoto, Thomas~F. Icard, Tianyi Zhang, Vishrav
  Chaudhary, William Wang, Xuechen Li, Yifan Mai, Yuhui Zhang, and Yuta
  Koreeda. 2022{\natexlab{b}}.
\newblock \href {https://arxiv.org/abs/2211.09110} {Holistic evaluation of
  language models}.
\newblock \emph{ArXiv}, abs/2211.09110.

\bibitem[{Liao et~al.(2021)Liao, Taori, Raji, and Schmidt}]{liao2021are}
Thomas Liao, Rohan Taori, Inioluwa~Deborah Raji, and Ludwig Schmidt. 2021.
\newblock \href {https://openreview.net/forum?id=mPducS1MsEK} {Are we learning
  yet? a meta review of evaluation failures across machine learning}.
\newblock In \emph{Thirty-fifth Conference on Neural Information Processing
  Systems Datasets and Benchmarks Track (Round 2)}.

\bibitem[{Liberman(2005)}]{liberman2005joshi}
Mark Liberman. 2005.
\newblock \href
  {http://itre.cis.upenn.edu/~myl/languagelog/archives/001989.html} {{Franklin
  Medal to Aravind Joshi}}.
\newblock \emph{Language Log}.

\bibitem[{Linzen(2020)}]{linzen2020accelerate}
Tal Linzen. 2020.
\newblock \href {https://doi.org/10.18653/v1/2020.acl-main.465} {How can we
  accelerate progress towards human-like linguistic generalization?}
\newblock In \emph{Proceedings of the 58th Annual Meeting of the Association
  for Computational Linguistics}, pages 5210--5217, Online. Association for
  Computational Linguistics.

\bibitem[{Liu et~al.(2021)Liu, Lee, Jia, and Liang}]{liu2021small}
Nelson~F. Liu, Tony Lee, Robin Jia, and Percy Liang. 2021.
\newblock \href {https://arxiv.org/abs/2102.01065} {Can small and synthetic
  benchmarks drive modeling innovation? a retrospective study of question
  answering modeling approaches}.
\newblock ArXiv:2102.01065.

\bibitem[{Miltsakaki et~al.(2004)Miltsakaki, Prasad, Joshi, and
  Webber}]{miltsakaki2004pdtb}
Eleni Miltsakaki, Rashmi Prasad, Aravind Joshi, and Bonnie Webber. 2004.
\newblock \href {http://www.lrec-conf.org/proceedings/lrec2004/pdf/618.pdf}
  {The {P}enn {D}iscourse {T}reebank}.
\newblock In \emph{Proceedings of the Fourth International Conference on
  Language Resources and Evaluation ({LREC}{'}04)}, Lisbon, Portugal. European
  Language Resources Association (ELRA).

\bibitem[{Nie et~al.(2020)Nie, Williams, Dinan, Bansal, Weston, and
  Kiela}]{nie2020anli}
Yixin Nie, Adina Williams, Emily Dinan, Mohit Bansal, Jason Weston, and Douwe
  Kiela. 2020.
\newblock \href {https://doi.org/10.18653/v1/2020.acl-main.441} {Adversarial
  {NLI}: A new benchmark for natural language understanding}.
\newblock In \emph{Proceedings of the 58th Annual Meeting of the Association
  for Computational Linguistics}, pages 4885--4901, Online. Association for
  Computational Linguistics.

\bibitem[{Nissim et~al.(2017)Nissim, Abzianidze, Evang, van~der Goot, Haagsma,
  Plank, and Wieling}]{nissim2017sharing}
Malvina Nissim, Lasha Abzianidze, Kilian Evang, Rob van~der Goot, Hessel
  Haagsma, Barbara Plank, and Martijn Wieling. 2017.
\newblock \href {https://doi.org/10.1162/COLI_a_00304} {Sharing is caring: The
  future of shared tasks}.
\newblock \emph{Computational Linguistics}, 43(4):897--904.

\bibitem[{Nivre et~al.(2016)Nivre, de~Marneffe, Ginter, Goldberg, Haji{\v{c}},
  Manning, McDonald, Petrov, Pyysalo, Silveira, Tsarfaty, and
  Zeman}]{nivre2016universal}
Joakim Nivre, Marie-Catherine de~Marneffe, Filip Ginter, Yoav Goldberg, Jan
  Haji{\v{c}}, Christopher~D. Manning, Ryan McDonald, Slav Petrov, Sampo
  Pyysalo, Natalia Silveira, Reut Tsarfaty, and Daniel Zeman. 2016.
\newblock \href {https://aclanthology.org/L16-1262} {{U}niversal {D}ependencies
  v1: A multilingual treebank collection}.
\newblock In \emph{Proceedings of the Tenth International Conference on
  Language Resources and Evaluation ({LREC}'16)}, pages 1659--1666,
  Portoro{\v{z}}, Slovenia. European Language Resources Association (ELRA).

\bibitem[{Parra~Escart{\'\i}n et~al.(2017)Parra~Escart{\'\i}n, Reijers, Lynn,
  Moorkens, Way, and Liu}]{parra2017shared}
Carla Parra~Escart{\'\i}n, Wessel Reijers, Teresa Lynn, Joss Moorkens, Andy
  Way, and Chao-Hong Liu. 2017.
\newblock \href {https://doi.org/10.18653/v1/W17-1608} {Ethical considerations
  in {NLP} shared tasks}.
\newblock In \emph{Proceedings of the First {ACL} Workshop on Ethics in Natural
  Language Processing}, pages 66--73, Valencia, Spain. Association for
  Computational Linguistics.

\bibitem[{Paullada et~al.(2021)Paullada, Raji, Bender, Denton, and
  Hanna}]{paullada2021data}
Amandalynne Paullada, Inioluwa~Deborah Raji, Emily~M. Bender, Emily Denton, and
  Alex Hanna. 2021.
\newblock \href {https://doi.org/https://doi.org/10.1016/j.patter.2021.100336}
  {Data and its (dis)contents: A survey of dataset development and use in
  machine learning research}.
\newblock \emph{Patterns}, 2(11):100336.

\bibitem[{Peng et~al.(2021)Peng, Mathur, and Narayanan}]{peng2021mitigating}
Kenneth~L Peng, Arunesh Mathur, and Arvind Narayanan. 2021.
\newblock \href {https://openreview.net/forum?id=KGeAHDH4njY} {Mitigating
  dataset harms requires stewardship: Lessons from 1000 papers}.
\newblock In \emph{Thirty-fifth Conference on Neural Information Processing
  Systems Datasets and Benchmarks Track (Round 2)}.

\bibitem[{Prasad et~al.(2008)Prasad, Dinesh, Lee, Miltsakaki, Robaldo, Joshi,
  and Webber}]{prasad2008pdtb2}
Rashmi Prasad, Nikhil Dinesh, Alan Lee, Eleni Miltsakaki, Livio Robaldo,
  Aravind Joshi, and Bonnie Webber. 2008.
\newblock \href
  {http://www.lrec-conf.org/proceedings/lrec2008/pdf/754_paper.pdf} {The {P}enn
  {D}iscourse {T}ree{B}ank 2.0.}
\newblock In \emph{Proceedings of the Sixth International Conference on
  Language Resources and Evaluation ({LREC}'08)}, Marrakech, Morocco. European
  Language Resources Association (ELRA).

\bibitem[{Prasad et~al.(2014)Prasad, Webber, and Joshi}]{prasad2014reflections}
Rashmi Prasad, Bonnie Webber, and Aravind Joshi. 2014.
\newblock \href {https://doi.org/10.1162/COLI_a_00204} {Reflections on the
  {P}enn {D}iscourse {T}ree{B}ank, comparable corpora, and complementary
  annotation}.
\newblock \emph{Computational Linguistics}, 40(4):921--950.

\bibitem[{Raji and Buolamwini(2019)}]{raji2019actionable}
Inioluwa~Deborah Raji and Joy Buolamwini. 2019.
\newblock \href {https://doi.org/10.1145/3306618.3314244} {Actionable auditing:
  Investigating the impact of publicly naming biased performance results of
  commercial ai products}.
\newblock In \emph{Proceedings of the 2019 AAAI/ACM Conference on AI, Ethics,
  and Society}, AIES '19, page 429–435, New York, NY, USA. Association for
  Computing Machinery.

\bibitem[{Raji et~al.(2021)Raji, Denton, Bender, Hanna, and
  Paullada}]{raji2021ai}
Inioluwa~Deborah Raji, Emily Denton, Emily~M. Bender, Alex Hanna, and
  Amandalynne Paullada. 2021.
\newblock \href {https://openreview.net/forum?id=j6NxpQbREA1} {{AI} and the
  everything in the whole wide world benchmark}.
\newblock In \emph{Thirty-fifth Conference on Neural Information Processing
  Systems Datasets and Benchmarks Track (Round 2)}.

\bibitem[{Raji et~al.(2020)Raji, Gebru, Mitchell, Buolamwini, Lee, and
  Denton}]{raji2020saving}
Inioluwa~Deborah Raji, Timnit Gebru, Margaret Mitchell, Joy Buolamwini,
  Joonseok Lee, and Emily Denton. 2020.
\newblock \href {https://doi.org/10.1145/3375627.3375820} {Saving face:
  Investigating the ethical concerns of facial recognition auditing}.
\newblock In \emph{Proceedings of the AAAI/ACM Conference on AI, Ethics, and
  Society}, AIES '20, page 145–151, New York, NY, USA. Association for
  Computing Machinery.

\bibitem[{Rajpurkar et~al.(2016)Rajpurkar, Zhang, Lopyrev, and
  Liang}]{rajpurkar2016squad}
Pranav Rajpurkar, Jian Zhang, Konstantin Lopyrev, and Percy Liang. 2016.
\newblock \href {https://doi.org/10.18653/v1/D16-1264} {{SQ}u{AD}: 100,000+
  questions for machine comprehension of text}.
\newblock In \emph{Proceedings of the 2016 Conference on Empirical Methods in
  Natural Language Processing}, pages 2383--2392, Austin, Texas. Association
  for Computational Linguistics.

\bibitem[{Rogers(2020)}]{rogers2020resource}
Anna Rogers. 2020.
\newblock \href {https://hackingsemantics.xyz/2020/reviewing-data/} {Peer
  review in nlp: resource papers}.

\bibitem[{Scheuerman et~al.(2021)Scheuerman, Hanna, and
  Denton}]{scheuerman2021politics}
Morgan~Klaus Scheuerman, Alex Hanna, and Emily Denton. 2021.
\newblock \href {https://doi.org/10.1145/3476058} {Do datasets have politics?
  disciplinary values in computer vision dataset development}.
\newblock \emph{Proc. ACM Hum.-Comput. Interact.}, 5(CSCW2).

\bibitem[{Sp\"arck~Jones and Galliers(1995)}]{jones1995evaluating}
Karen Sp\"arck~Jones and Julia~R. Galliers. 1995.
\newblock \emph{Evaluating Natural Language Processing Systems: An Analysis and
  Review}.
\newblock Number 1083 in Lecture Notes in Computer Science. Springer Verlag,
  Berlin.

\bibitem[{Srivastava et~al.(2022)Srivastava, Rastogi, Rao, Shoeb, Abid, Fisch,
  Brown, Santoro, Gupta, Garriga-Alonso, Kluska, Lewkowycz, Agarwal, Power,
  Ray, Warstadt, Kocurek, Safaya, Tazarv, Xiang, Parrish, Nie, Hussain, Askell,
  Dsouza, Rahane, Iyer, Andreassen, Santilli, Stuhlmuller, Dai, La, Lampinen,
  Zou, Jiang, Chen, Vuong, Gupta, Gottardi, Norelli, Venkatesh, Gholamidavoodi,
  Tabassum, Menezes, Kirubarajan, Mullokandov, Sabharwal, Herrick, Efrat,
  Erdem, Karakacs, Roberts, Loe, Zoph, Bojanowski, Ozyurt, Hedayatnia,
  Neyshabur, Inden, Stein, Ekmekci, Lin, Howald, Diao, Dour, Stinson, Argueta,
  Ram'irez, Singh, Rathkopf, Meng, Baral, Wu, Callison-Burch, Waites, Voigt,
  Manning, Potts, Ramirez, Rivera, Siro, Raffel, Ashcraft, Garbacea, Sileo,
  Garrette, Hendrycks, Kilman, Roth, Freeman, Khashabi, Levy, Gonz'alez,
  Hernandez, Chen, Ippolito, Gilboa, Dohan, Drakard, Jurgens, Datta, Ganguli,
  Emelin, Kleyko, Yuret, Chen, Tam, Hupkes, Misra, Buzan, Mollo, Yang, Lee,
  Shutova, Cubuk, Segal, Hagerman, Barnes, Donoway, Pavlick, Rodol{\`a}, Lam,
  Chu, Tang, Erdem, Chang, Chi, Dyer, Jerzak, Kim, Manyasi, Zheltonozhskii,
  Xia, Siar, Mart'inez-Plumed, Happ'e, Chollet, Rong, Mishra, Winata, de~Melo,
  Kruszewski, Parascandolo, Mariani, Wang, Jaimovitch-L'opez, Betz, Gur-Ari,
  Galijasevic, Kim, Rashkin, Hajishirzi, Mehta, Bogar, Shevlin, Sch{\"u}tze,
  Yakura, Zhang, Wong, Ng, Noble, Jumelet, Geissinger, Kernion, Hilton, Lee,
  Fisac, Simon, Koppel, Zheng, Zou, Koco'n, Thompson, Kaplan, Radom,
  Sohl-Dickstein, Phang, Wei, Yosinski, Novikova, Bosscher, Marsh, Kim, Taal,
  Engel, Alabi, Xu, Song, Tang, Waweru, Burden, Miller, Balis, Berant,
  Frohberg, Rozen, Hern{\'a}ndez-Orallo, Boudeman, Jones, Tenenbaum, Rule,
  Chua, Kanclerz, Livescu, Krauth, Gopalakrishnan, Ignatyeva, Markert, Dhole,
  Gimpel, Omondi, Mathewson, Chiafullo, Shkaruta, Shridhar, McDonell,
  Richardson, Reynolds, Gao, Zhang, Dugan, Qin, Contreras-Ochando, Morency,
  Moschella, Lam, Noble, Schmidt, He, Col'on, Metz, cSenel, Bosma, Sap, ter
  Hoeve, Andrea, Farooqi, Faruqui, Mazeika, Baturan, Marelli, Maru, Quintana,
  Tolkiehn, Giulianelli, Lewis, Potthast, Leavitt, Hagen, Schubert,
  Baitemirova, Arnaud, McElrath, Yee, Cohen, Gu, Ivanitskiy, Starritt, Strube,
  Swkedrowski, Bevilacqua, Yasunaga, Kale, Cain, Xu, Suzgun, Tiwari, Bansal,
  Aminnaseri, Geva, Gheini, MukundVarma, Peng, Chi, Lee, Krakover, Cameron,
  Roberts, Doiron, Nangia, Deckers, Muennighoff, Keskar, Iyer, Constant,
  Fiedel, Wen, Zhang, Agha, Elbaghdadi, Levy, Evans, Casares, Doshi, Fung,
  Liang, Vicol, Alipoormolabashi, Liao, Liang, Chang, Eckersley, Htut, Hwang,
  Milkowski, Patil, Pezeshkpour, Oli, Mei, LYU, Chen, Banjade, Rudolph,
  Gabriel, Habacker, Delgado, Milli{\`e}re, Garg, Barnes, Saurous, Arakawa,
  Raymaekers, Frank, Sikand, Novak, Sitelew, Bras, Liu, Jacobs, Zhang,
  Salakhutdinov, Chi, Lee, Stovall, Teehan, Yang, Singh, Mohammad, Anand,
  Dillavou, Shleifer, Wiseman, Gruetter, Bowman, Schoenholz, Han, Kwatra, Rous,
  Ghazarian, Ghosh, Casey, Bischoff, Gehrmann, Schuster, Sadeghi, Hamdan, Zhou,
  Srivastava, Shi, Singh, Asaadi, Gu, Pachchigar, Toshniwal, Upadhyay, Debnath,
  Shakeri, Thormeyer, Melzi, Reddy, Makini, hwan Lee, Torene, Hatwar, Dehaene,
  Divic, Ermon, Biderman, Lin, Prasad, Piantadosi, Shieber, Misherghi,
  Kiritchenko, Mishra, Linzen, Schuster, Li, Yu, Ali, Hashimoto, Wu, Desbordes,
  Rothschild, Phan, Wang, Nkinyili, Schick, Kornev, Telleen-Lawton, Tunduny,
  Gerstenberg, Chang, Neeraj, Khot, Shultz, Shaham, Misra, Demberg, Nyamai,
  Raunak, Ramasesh, Prabhu, Padmakumar, Srikumar, Fedus, Saunders, Zhang,
  Vossen, Ren, Tong, Wu, Shen, Yaghoobzadeh, Lakretz, Song, Bahri, Choi, Yang,
  Hao, Chen, Belinkov, Hou, Hou, Bai, Seid, Xinran, Zhao, Wang, Wang, Wang, Wu,
  Singh, and Shaham}]{srivastava2022bigbench}
Aarohi Srivastava, Abhinav Rastogi, Abhishek~B Rao, Abu Awal~Md Shoeb, Abubakar
  Abid, Adam Fisch, Adam~R. Brown, Adam Santoro, Aditya Gupta, Adri{\`a}
  Garriga-Alonso, Agnieszka Kluska, Aitor Lewkowycz, Akshat Agarwal, Alethea
  Power, Alex Ray, Alex Warstadt, Alexander~W. Kocurek, Ali Safaya, Ali Tazarv,
  Alice Xiang, Alicia Parrish, Allen Nie, Aman Hussain, Amanda Askell, Amanda
  Dsouza, Ameet~Annasaheb Rahane, Anantharaman~S. Iyer, Anders Andreassen,
  Andrea Santilli, Andreas Stuhlmuller, Andrew~M. Dai, Andrew~D. La,
  Andrew~Kyle Lampinen, Andy Zou, Angela Jiang, Angelica Chen, Anh Vuong,
  Animesh Gupta, Anna Gottardi, Antonio Norelli, Anu Venkatesh, Arash
  Gholamidavoodi, Arfa Tabassum, Arul Menezes, Arun Kirubarajan, Asher
  Mullokandov, Ashish Sabharwal, Austin Herrick, Avia Efrat, Aykut Erdem, Ayla
  Karakacs, Bridget~R. Roberts, Bao~Sheng Loe, Barret Zoph, Bartlomiej
  Bojanowski, Batuhan Ozyurt, Behnam Hedayatnia, Behnam Neyshabur, Benjamin
  Inden, Benno Stein, Berk Ekmekci, Bill~Yuchen Lin, Blake~Stephen Howald,
  Cameron Diao, Cameron Dour, Catherine Stinson, Cedrick Argueta, C'esar~Ferri
  Ram'irez, Chandan Singh, Charles Rathkopf, Chenlin Meng, Chitta Baral, Chiyu
  Wu, Chris Callison-Burch, Chris Waites, Christian Voigt, Christopher~D.
  Manning, Christopher Potts, Cindy~Tatiana Ramirez, Clara Rivera, Clemencia
  Siro, Colin Raffel, Courtney Ashcraft, Cristina Garbacea, Damien Sileo,
  Daniel~H Garrette, Dan Hendrycks, Dan Kilman, Dan Roth, Daniel Freeman,
  Daniel Khashabi, Daniel Levy, Daniel Gonz'alez, Danny Hernandez, Danqi Chen,
  Daphne Ippolito, Dar Gilboa, David Dohan, D.~Drakard, David Jurgens,
  Debajyoti Datta, Deep Ganguli, Denis Emelin, Denis Kleyko, Deniz Yuret, Derek
  Chen, Derek Tam, Dieuwke Hupkes, Diganta Misra, Dilyar Buzan, Dimitri~Coelho
  Mollo, Diyi Yang, Dong-Ho Lee, Ekaterina Shutova, Ekin~Dogus Cubuk, Elad
  Segal, Eleanor Hagerman, Elizabeth Barnes, Elizabeth~P. Donoway, Ellie
  Pavlick, Emanuele Rodol{\`a}, Emma~FC Lam, Eric Chu, Eric Tang, Erkut Erdem,
  Ernie Chang, Ethan~A. Chi, Ethan Dyer, Ethan Jerzak, Ethan Kim, Eunice~Engefu
  Manyasi, Evgenii Zheltonozhskii, Fan Xia, Fatemeh Siar, Fernando
  Mart'inez-Plumed, Francesca Happ'e, François Chollet, Frieda Rong, Gaurav
  Mishra, Genta~Indra Winata, Gerard de~Melo, Germ{\'a}n Kruszewski,
  Giambattista Parascandolo, Giorgio Mariani, Gloria Wang, Gonzalo
  Jaimovitch-L'opez, Gregor Betz, Guy Gur-Ari, Hana Galijasevic, Han~Sol Kim,
  Hannah Rashkin, Hanna Hajishirzi, Harsh Mehta, Hayden Bogar, Henry Shevlin,
  Hinrich Sch{\"u}tze, Hiromu Yakura, Hongming Zhang, Hubert Wong, Ian Aik-Soon
  Ng, Isaac Noble, Jaap Jumelet, Jack Geissinger, John Kernion, Jacob Hilton,
  Jaehoon Lee, Jaime~Fern{\'a}ndez Fisac, J.~Brooker Simon, James Koppel, James
  Zheng, James Zou, Jan Koco'n, Jana Thompson, Jared Kaplan, Jarema Radom,
  Jascha~Narain Sohl-Dickstein, Jason Phang, Jason Wei, Jason Yosinski,
  Jekaterina Novikova, Jelle Bosscher, Jenni Marsh, Jeremy Kim, Jeroen Taal,
  Jesse Engel, Jesujoba~Oluwadara Alabi, Jiacheng Xu, Jiaming Song, Jillian
  Tang, Jane~W Waweru, John Burden, John Miller, John~U. Balis, Jonathan
  Berant, Jorg Frohberg, Jos Rozen, Jos{\'e} Hern{\'a}ndez-Orallo, Joseph
  Boudeman, Joseph Jones, Joshua~B. Tenenbaum, Joshua~S. Rule, Joyce Chua,
  Kamil Kanclerz, Karen Livescu, Karl Krauth, Karthik Gopalakrishnan, Katerina
  Ignatyeva, Katja Markert, Kaustubh~D. Dhole, Kevin Gimpel, Kevin~Ochieng’
  Omondi, Kory~Wallace Mathewson, Kristen Chiafullo, Ksenia Shkaruta, Kumar
  Shridhar, Kyle McDonell, Kyle Richardson, Laria Reynolds, Leo Gao, Li~Zhang,
  Liam Dugan, Lianhui Qin, Lidia Contreras-Ochando, Louis-Philippe Morency,
  Luca Moschella, Luca Lam, Lucy Noble, Ludwig Schmidt, Luheng He,
  Luis~Oliveros Col'on, Luke Metz, Lutfi~Kerem cSenel, Maarten Bosma, Maarten
  Sap, Maartje ter Hoeve, Madotto Andrea, Maheen~Saleem Farooqi, Manaal
  Faruqui, Mantas Mazeika, Marco Baturan, Marco Marelli, Marco Maru,
  M~Quintana, Marie Tolkiehn, Mario Giulianelli, Martha Lewis, Martin Potthast,
  Matthew Leavitt, Matthias Hagen, M'aty'as Schubert, Medina Baitemirova,
  Melissa Arnaud, Melvin~Andrew McElrath, Michael~A. Yee, Michael Cohen, Mi~Gu,
  Michael~I. Ivanitskiy, Michael Starritt, Michael Strube, Michal Swkedrowski,
  Michele Bevilacqua, Michihiro Yasunaga, Mihir Kale, Mike Cain, Mimee Xu,
  Mirac Suzgun, Monica Tiwari, Mohit Bansal, Moin Aminnaseri, Mor Geva, Mozhdeh
  Gheini, T~MukundVarma, Nanyun Peng, Nathan Chi, Nayeon Lee, Neta Gur-Ari
  Krakover, Nicholas Cameron, Nicholas~S. Roberts, Nicholas Doiron, Nikita
  Nangia, Niklas Deckers, Niklas Muennighoff, Nitish~Shirish Keskar, Niveditha
  Iyer, Noah Constant, Noah Fiedel, Nuan Wen, Oliver Zhang, Omar Agha, Omar
  Elbaghdadi, Omer Levy, Owain Evans, Pablo Antonio~Moreno Casares, Parth
  Doshi, Pascale Fung, Paul~Pu Liang, Paul Vicol, Pegah Alipoormolabashi,
  Peiyuan Liao, Percy Liang, Peter~W. Chang, Peter Eckersley, Phu~Mon Htut,
  Pi-Bei Hwang, P.~Milkowski, Piyush~S. Patil, Pouya Pezeshkpour, Priti Oli,
  Qiaozhu Mei, QING LYU, Qinlang Chen, Rabin Banjade, Rachel~Etta Rudolph,
  Raefer Gabriel, Rahel Habacker, Ram'on~Risco Delgado, Rapha{\"e}l
  Milli{\`e}re, Rhythm Garg, Richard Barnes, Rif~A. Saurous, Riku Arakawa,
  Robbe Raymaekers, Robert Frank, Rohan Sikand, Roman Novak, Roman Sitelew,
  Ronan~Le Bras, Rosanne Liu, Rowan Jacobs, Rui Zhang, Ruslan Salakhutdinov,
  Ryan Chi, Ryan Lee, Ryan Stovall, Ryan Teehan, Rylan Yang, Sahib~J. Singh,
  Saif~M. Mohammad, Sajant Anand, Sam Dillavou, Sam Shleifer, Sam Wiseman,
  Samuel Gruetter, Sam Bowman, Samuel~S. Schoenholz, Sanghyun Han, Sanjeev
  Kwatra, Sarah~A. Rous, Sarik Ghazarian, Sayan Ghosh, Sean Casey, Sebastian
  Bischoff, Sebastian Gehrmann, Sebastian Schuster, Sepideh Sadeghi, Shadi~S.
  Hamdan, Sharon Zhou, Shashank Srivastava, Sherry Shi, Shikhar Singh, Shima
  Asaadi, Shixiang~Shane Gu, Shubh Pachchigar, Shubham Toshniwal, Shyam
  Upadhyay, Shyamolima Debnath, Siamak Shakeri, Simon Thormeyer, Simone Melzi,
  Siva Reddy, Sneha~Priscilla Makini, Soo hwan Lee, Spencer~Bradley Torene,
  Sriharsha Hatwar, Stanislas Dehaene, Stefan Divic, Stefano Ermon, Stella~Rose
  Biderman, Stephanie~C. Lin, Stephen Prasad, Steven~T. Piantadosi, Stuart~M.
  Shieber, Summer Misherghi, Svetlana Kiritchenko, Swaroop Mishra, Tal Linzen,
  Tal Schuster, Tao Li, Tao Yu, Tariq~A. Ali, Tatsuo Hashimoto, Te-Lin Wu, Theo
  Desbordes, Theodore Rothschild, Thomas Phan, Tianle Wang, Tiberius Nkinyili,
  Timo Schick, T.~N. Kornev, Timothy Telleen-Lawton, Titus Tunduny, Tobias
  Gerstenberg, Trenton Chang, Trishala Neeraj, Tushar Khot, Tyler~O’Brien
  Shultz, Uri Shaham, Vedant Misra, Vera Demberg, Victoria Nyamai, Vikas
  Raunak, Vinay~Venkatesh Ramasesh, Vinay~Uday Prabhu, Vishakh Padmakumar,
  Vivek Srikumar, William Fedus, William Saunders, William Zhang, W~Vossen,
  Xiang Ren, Xiaoyu~F Tong, Xinyi Wu, Xudong Shen, Yadollah Yaghoobzadeh, Yair
  Lakretz, Yang Song, Yasaman Bahri, Ye~Ji Choi, Yichi Yang, Yiding Hao, Yifu
  Chen, Yonatan Belinkov, Yu~Hou, Yu~Hou, Yushi Bai, Zachary Seid, Zhao Xinran,
  Zhuoye Zhao, Zi~Fu Wang, Zijie~J. Wang, Zirui Wang, Ziyi Wu, Sahib Singh, and
  Uri Shaham. 2022.
\newblock \href {https://arxiv.org/abs/2206.04615} {Beyond the imitation game:
  Quantifying and extrapolating the capabilities of language models}.
\newblock \emph{ArXiv}, abs/2206.04615.

\bibitem[{Vaswani et~al.(2017)Vaswani, Shazeer, Parmar, Uszkoreit, Jones,
  Gomez, Kaiser, and Polosukhin}]{vaswani2017transformers}
Ashish Vaswani, Noam Shazeer, Niki Parmar, Jakob Uszkoreit, Llion Jones,
  Aidan~N Gomez, \L~ukasz Kaiser, and Illia Polosukhin. 2017.
\newblock \href
  {https://proceedings.neurips.cc/paper/2017/file/3f5ee243547dee91fbd053c1c4a845aa-Paper.pdf}
  {Attention is all you need}.
\newblock In \emph{Advances in Neural Information Processing Systems},
  volume~30. Curran Associates, Inc.

\bibitem[{von Werra et~al.(2022)von Werra, Tunstall, Thakur, Luccioni, Thrush,
  Piktus, Marty, Rajani, Mustar, Ngo, Sanseviero, vSavsko, Villanova, Lhoest,
  Chaumond, Mitchell, Rush, Wolf, and Kiela}]{werra2022evaluate}
Leandro von Werra, Lewis Tunstall, Abhishek Thakur, Alexandra~Sasha Luccioni,
  Tristan Thrush, Aleksandra Piktus, Felix Marty, Nazneen Rajani, Victor
  Mustar, Helen Ngo, Omar Sanseviero, Mario vSavsko, Albert Villanova, Quentin
  Lhoest, Julien Chaumond, Margaret Mitchell, Alexander~M. Rush, Thomas Wolf,
  and Douwe Kiela. 2022.
\newblock Evaluate\&evaluation on the hub: Better best practices for data and
  model measurements.

\bibitem[{Wang et~al.(2019{\natexlab{a}})Wang, Pruksachatkun, Nangia, Singh,
  Michael, Hill, Levy, and Bowman}]{wang2019superglue}
Alex Wang, Yada Pruksachatkun, Nikita Nangia, Amanpreet Singh, Julian Michael,
  Felix Hill, Omer Levy, and Samuel Bowman. 2019{\natexlab{a}}.
\newblock \href
  {https://proceedings.neurips.cc/paper/2019/file/4496bf24afe7fab6f046bf4923da8de6-Paper.pdf}
  {Superglue: A stickier benchmark for general-purpose language understanding
  systems}.
\newblock In \emph{Advances in Neural Information Processing Systems},
  volume~32. Curran Associates, Inc.

\bibitem[{Wang et~al.(2019{\natexlab{b}})Wang, Singh, Michael, Hill, Levy, and
  Bowman}]{wang2019glue}
Alex Wang, Amanpreet Singh, Julian Michael, Felix Hill, Omer Levy, and
  Samuel~R. Bowman. 2019{\natexlab{b}}.
\newblock \href {https://openreview.net/forum?id=rJ4km2R5t7} {{GLUE}: A
  multi-task benchmark and analysis platform for natural language
  understanding}.
\newblock In \emph{International Conference on Learning Representations}.

\bibitem[{Webber(2018)}]{webber2018obituary}
Bonnie Webber. 2018.
\newblock \href {https://doi.org/10.1162/coli_a_00321} {{O}bituary: {A}ravind
  {K}. {J}oshi}.
\newblock \emph{Computational Linguistics}, 44(3):387--392.

\bibitem[{Wei et~al.(2022)Wei, Tay, Bommasani, Raffel, Zoph, Borgeaud,
  Yogatama, Bosma, Zhou, Metzler, Chi, Hashimoto, Vinyals, Liang, Dean, and
  Fedus}]{wei2022emergent}
Jason Wei, Yi~Tay, Rishi Bommasani, Colin Raffel, Barret Zoph, Sebastian
  Borgeaud, Dani Yogatama, Maarten Bosma, Denny Zhou, Donald Metzler, Ed~H.
  Chi, Tatsunori Hashimoto, Oriol Vinyals, Percy Liang, Jeff Dean, and William
  Fedus. 2022.
\newblock \href {https://openreview.net/forum?id=yzkSU5zdwD} {Emergent
  abilities of large language models}.
\newblock \emph{Transactions on Machine Learning Research}.
\newblock Survey Certification.

\bibitem[{{WELM}(2021)}]{welm2021welm}
{WELM}. 2021.
\newblock \href {https://welmworkshop.github.io/} {{Workshop on Enormous
  Language Models (WELM)}}.

\bibitem[{Welty et~al.(2019)Welty, Paritosh, and Aroyo}]{welty2019metrology}
Chris Welty, Praveen~K. Paritosh, and Lora Aroyo. 2019.
\newblock \href {https://arxiv.org/abs/1911.01875} {Metrology for ai: From
  benchmarks to instruments}.
\newblock \emph{ArXiv}, abs/1911.01875.

\bibitem[{Wolf et~al.(2020)Wolf, Debut, Sanh, Chaumond, Delangue, Moi, Cistac,
  Rault, Louf, Funtowicz, Davison, Shleifer, von Platen, Ma, Jernite, Plu, Xu,
  Le~Scao, Gugger, Drame, Lhoest, and Rush}]{wolf2020transformers}
Thomas Wolf, Lysandre Debut, Victor Sanh, Julien Chaumond, Clement Delangue,
  Anthony Moi, Pierric Cistac, Tim Rault, Remi Louf, Morgan Funtowicz, Joe
  Davison, Sam Shleifer, Patrick von Platen, Clara Ma, Yacine Jernite, Julien
  Plu, Canwen Xu, Teven Le~Scao, Sylvain Gugger, Mariama Drame, Quentin Lhoest,
  and Alexander Rush. 2020.
\newblock \href {https://doi.org/10.18653/v1/2020.emnlp-demos.6} {Transformers:
  State-of-the-art natural language processing}.
\newblock In \emph{Proceedings of the 2020 Conference on Empirical Methods in
  Natural Language Processing: System Demonstrations}, pages 38--45, Online.
  Association for Computational Linguistics.

\bibitem[{Yuan et~al.(2022)Yuan, He, Davis, Zhang, Dao, Chen, Liang, Re, and
  Zhang}]{yuan2022decentralized}
Binhang Yuan, Yongjun He, Jared~Quincy Davis, Tianyi Zhang, Tri Dao, Beidi
  Chen, Percy Liang, Christopher Re, and Ce~Zhang. 2022.
\newblock \href {https://openreview.net/forum?id=UHoGOaGjEq} {Decentralized
  training of foundation models in heterogeneous environments}.
\newblock In \emph{Advances in Neural Information Processing Systems}.

\end{thebibliography}
\bibliographystyle{acl_natbib}
\end{document}